\documentclass[letterpaper, 10pt, conference]{ieeeconf}
\IEEEoverridecommandlockouts
\let\labelindent\relax
\usepackage{enumitem}
\usepackage{cite}
\usepackage{amsmath,amssymb,amsfonts}
\usepackage{algorithmic}
\usepackage{graphicx}
\usepackage{textcomp}
\usepackage{xcolor}
\usepackage{comment}
\def\BibTeX{{\rm B\kern-.05em{\sc i\kern-.025em b}\kern-.08em
    T\kern-.1667em\lower.7ex\hbox{E}\kern-.125emX}}

\usepackage[nolist]{acronym}
\usepackage{todonotes}
\usepackage{skeldoc} 
\usepackage{hyperref}
\usepackage[nameinlink, capitalise]{cleveref}
\usepackage{layouts}
\usepackage{tikz}
\usepackage{pgfplots}
\crefformat{footnote}{#2\footnotemark[#1]#3}

\usepackage{siunitx}      
\usepackage{multirow}     
\usepackage{caption}      
\usepackage{booktabs}     
\usepackage{caption}      
\usepackage{siunitx}      
\usepackage{dblfloatfix}  

\usepackage{fancyhdr}   
\fancypagestyle{firstpage}{%
 \fancyhf{}%
 \fancyfoot[L]{%
  \small{ ~\copyright2024 IEEE. Personal use of this material is permitted.  Permission from IEEE must be obtained for all other uses, in any current or future media, including reprinting/republishing this material for advertising or promotional purposes, creating new collective works, for resale or redistribution to servers or lists, or reuse of any copyrighted component of this work in other works.}
 }%
}

\begin{document}
\newcommand{\mycomment}[1]{}
\title{\LARGE \bf Integrating End-to-End and Modular Driving Approaches for Online Corner Case Detection in Autonomous Driving}

\author{Gemb Kaljavesi$^{1}$, Xiyan Su$^{1}$, and Frank Diermeyer$^{1}$
\thanks{*This work was supported by the Federal Ministry of Education and Research of Germany~(BMBF) within the project ConnRAD~(FKZ~16KISR034), the project Wies'n Shuttle~(FKZ~03ZU1105AA) in the MCube cluster and through basic research funds from the Institute for Automotive Technology.}
\thanks{$^{1}$Gemb Kaljavesi, Xiyan Su and Frank Diermeyer are with the Institute of Automotive Technology at the Technical University of Munich (TUM), DE-85748 Garching, Germany. {\tt\small\{gemb.kaljavesi, xiyan.su, diermeyer\}@tum.de}}%
}

\mycomment{
\author{\IEEEauthorblockN{1\textsuperscript{st} Gemb Kaljavesi}
\IEEEauthorblockA{\textit{Institute of Automotive Technology } \\
\textit{Technical University of Munich}\\
Garching, Germany \\
gemb.kaljavesi@tum.de}
\and
\IEEEauthorblockN{2\textsuperscript{nd} Xiyan Su}
\IEEEauthorblockA{\textit{Institute of Automotive Technology } \\
\textit{Technical University of Munich}\\
Garching, Germany \\
xiyan.su@tum.de}
\and
\IEEEauthorblockN{6\textsuperscript{th} Frank Diermeyer}
\IEEEauthorblockA{\textit{Institute of Automotive Technology } \\
\textit{Technical University of Munich}\\
Garching, Germany \\
diermeyer@tum.de}
}
}
    \maketitle
    \thispagestyle{firstpage}

    \begin{abstract}
Online corner case detection is crucial for ensuring safety in autonomous driving vehicles. Current autonomous driving approaches can be categorized into modular approaches and end-to-end approaches. To leverage the advantages of both, we propose a method for online corner case detection that integrates an end-to-end approach into a modular system. The modular system takes over the primary driving task and the end-to-end network runs in parallel as a secondary one, the disagreement between the systems is then used for corner case detection. We implement this method on a real vehicle and evaluate it qualitatively. Our results demonstrate that end-to-end networks, known for their superior situational awareness, as secondary driving systems, can effectively contribute to corner case detection. These findings suggest that such an approach holds potential for enhancing the safety of autonomous vehicles.

\end{abstract}

    \section{Introduction}
\label{sec:introduction}

The advancement of autonomous vehicles (AVs) technology is rapidly approaching the point where their integration into mainstream transportation systems is becoming increasingly feasible. The impact of AVs on road safety has already been investigated in several studies, which were conducted through virtual traffic simulations \renewcommand \citepunct{, }\cite{tafidis19, morando18} and real-world evaluations using historical data \cite{combs19, utriainen20}. In general, these studies support the opinion that autonomous vehicles have the potential to improve road safety \cite{tafidis22}.

The research field of autonomous driving can be roughly divided into two categories: modular approaches and end-to-end approaches. Modular approaches divide the driving task into individual, self-contained parts that are interchangeable. In autonomous driving, the modules are typically categorized into mapping, localization, perception, planning, and control \cite{yurtsever20}. The modular approach is considered the conventional method and is currently more widespread in industry and research \cite{tampuu22}. On the other hand, end-to-end approaches encapsulate the entire driving task, or larger parts of it, as a single machine learning task. These approaches are becoming increasingly important \cite{chib23}; however, they are primarily trained and evaluated in simulations such as CARLA \cite{chib23, dosovitskiy, shao23a}. Particularly, end-to-end approaches with multi-modal input, which utilize more than just camera data, are rarely encountered in real vehicle applications. Both approaches have their respective advantages and disadvantages, which determine their suitability in specific situations \cite{tampuu22}.

However, both suffer from what many consider the major challenge: the long tail of rare events that these systems encounter \cite{jain21}. AVs operate in a complex environment and face intricate scenarios, not all of which can be accounted for during development \cite{schlatow17}. Such rare and unpredictable scenarios, encountered during real-world driving but infrequently observed in normal conditions, are often referred to as corner cases and their independent detection is essential for safe autonomous driving \cite{bogdoll22}. In recent years, several methods have been developed for this purpose. However, an online corner case detection using a combination of modular and end-to-end software approaches has not yet been explored.

\subsection{Related Work}
Modular approaches consist of a structured chain of modules that communicate with each other via predefined input and output data. Examples of current open-source modular approaches for autonomous driving are Autoware Universe \cite{git_autoware} and Apollo \cite{git_apollo}, both of which are already being deployed on real vehicles. One advantage of modular approaches is their interpretability. In the event of a malfunction, it is possible to trace the point at which the error occurred or the interactions within the system that led to this error \cite{betz19, plaza}. Moreover, even in successful missions, the behavior and reasoning of the system can be traced \cite{betz19, plaza}. In addition to that the clearly defined intermediate representations and deterministic rules ensure predictable behavior in autonomous-driving systems based on modular pipelines, given the strict interdependencies among subsystems \cite{tampuu22}. Building upon this fact, the approval of such a system is also more feasible \cite{stahl}. However, disadvantages arise from the fact that inputs and outputs are predefined, which may not be optimal for certain situations and limit the ability to articulate the finer details of the driving situations encountered \cite{tampuu22, jain21}. Additionally, the predefined outputs lead to a significant loss of information from the raw sensor data. For instance, in typical camera-based object detection, only the bounding boxes of the detected objects are retained, resulting in the loss of information from the remaining pixels \cite{gruyer17}.

Recent studies have increasingly focused on end-to-end approaches, particularly facilitated by the CARLA simulator and its associated leaderboard \cite{dosovitskiy, shao23a, shao22, chen22}. End-to-end systems consolidate the tasks of individual modules into a single neural network, optimized for one primary objective: the driving task \cite{chen24}. With undefined inputs and outputs, these systems can implicitly learn to utilize information \cite{tampuu22}. Moreover, they allow for full utilization of sensor data, without intermediate results as in the modular approach, enabling global reasoning within existing scenarios \cite{shao23a}. Despite progress, the interpretability of such approaches remains a major challenge \cite{shao22, chen20}. Moreover, additional effort is necessary to achieve a smooth trajectory and continuous speed, which can also be realised by the vehicle \cite{tampuu22, hecker2019learning}. Furthermore, obtaining approval for such systems in public road traffic presents an additional obstacle, which could currently only be addressed by additional rule-based approaches \cite{stahl, stahl21a}.

Corner cases can be categorized by their abstraction level \cite{breitenstein20}, but no metric or universally accepted definition exists \cite{bolte19, heidecker21}. While numerous works have focused on detecting corner cases at lower abstraction levels, such as the domain, object or scene level, detecting corner cases at the scenario level is much more complex \cite{bogdoll22}. This requires taking into account patterns with the addition of temporal context. The ISO/PAS 21448 SOTIF~\cite{isopas_21448_iso_2022} provides a defintion for further specifying corner cases on scenario level based on their hazardousness and novelty.

In recent years, several approaches have been published that enable online corner case detection at the scenario level. Approaches such as those in \cite{stocco20, hussain22} attempt to reconstruct sensor data from recent time steps using auto-encoders, primarily utilizing camera images. In the subsequent time step, the reconstruction undergoes verification, and depending on its quality, a corner case is identified. Research efforts, such as those by \cite{kuhn20, kuhn22} endeavor to predict online corner cases by learning from so-called disengagement scenarios, which entail recorded data from instances where a safety driver had to intervene. In \cite{fridman18, hanselaar24b}, corner case detection is achieved through several end-to-end systems for autonomous driving, characterized by a heterogeneous network architecture and concurrent operation. An end-to-end system serves as the primary system responsible for executing the driving task. The redundant system is present as a safety measure, detecting corner cases through deviations from the first system. However, such approaches also suffer from the same disadvantages of a single end-to-end system. So far, there is no approach that attempts to combine the advantages of the modular system with those of the end-to-end system for corner case detection. 

\subsection{Contribution}
This work introduces a novel concept: an online corner case detection, facilitated by the combination of end-to-end and modular autonomous software. This integration aims to harness the benefits offered by both approaches. Additionally, as part of this work, the concept is implemented and qualitatively evaluated using a research vehicle. For this purpose, the end-to-end network is initially trained in simulation and subsequently with real data. For evaluation, the trained network is then deployed on a research vehicle in parallel with a modular approach. 

The reminder of this work is structured as follows. \cref{sec:method} introduces the proposed methodology for online corner case detection. \cref{sec:experiment_design} provides a detailed explanation of the experimental implementation of the methodology, including the test setup and deployment on the research vehicle. \Cref{sec:results} presents the results, which are discussed in \cref{sec:discussion}, while \cref{sec:conclusion} draws conclusions from these results and outlines directions for future research.
    \section{Proposed Online Corner Case Detection}
\label{sec:method}

\begin{figure}[b]
\includegraphics[width=\columnwidth]{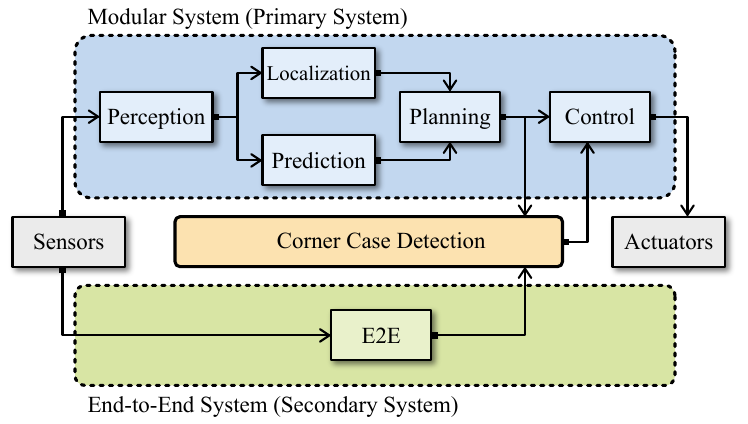}
\caption{Overview of the proposed method for online corner case detection through the integration of end-to-end and modular software. }
\label{fig:method}
\end{figure}%

The proposed methodology is depicted in \cref{fig:method}. Generally, the corner case detection is realized through the integration of an end-to-end system and a modular system. Within this framework, a divergence observed between the two systems is presumed to signify a corner case. In line with the methodology outlined in \cite{fridman18}, a division exists between a primary and a secondary system. The primary system assumes control over the driving task and is capable of issuing both lateral and longitudinal commands to the vehicle. In contrast, the secondary system lacks direct access to the vehicle's actuators and is only able to influence the longitudinal behavior of the vehicle. The modular system is selected as the primary system in the proposed method. This decision is based on the higher interpretability, potential predictability, and possible certification offered by the modular system, in comparison to the end-to-end systems already discussed in \cref{sec:introduction}. The end-to-end system, which offers superior global reasoning by utilizing the entirety of the data and holistic optimization for the driving function, provides a more comprehensive assessment of the current scenario. It is designated as the secondary system, implying it lacks direct control over the vehicle. Incorporating the end-to-end network as a secondary system introduces two new properties that can be considered by the network architecture:
\begin{enumerate}
    \item The network is not required to undertake the entire driving task. \label{point:task}
    \item The network is not obligated to perform its task in real-time at high frequency. \label{point:realtime}
\end{enumerate}

\begin{figure*}[b]
\includegraphics[width=\textwidth]{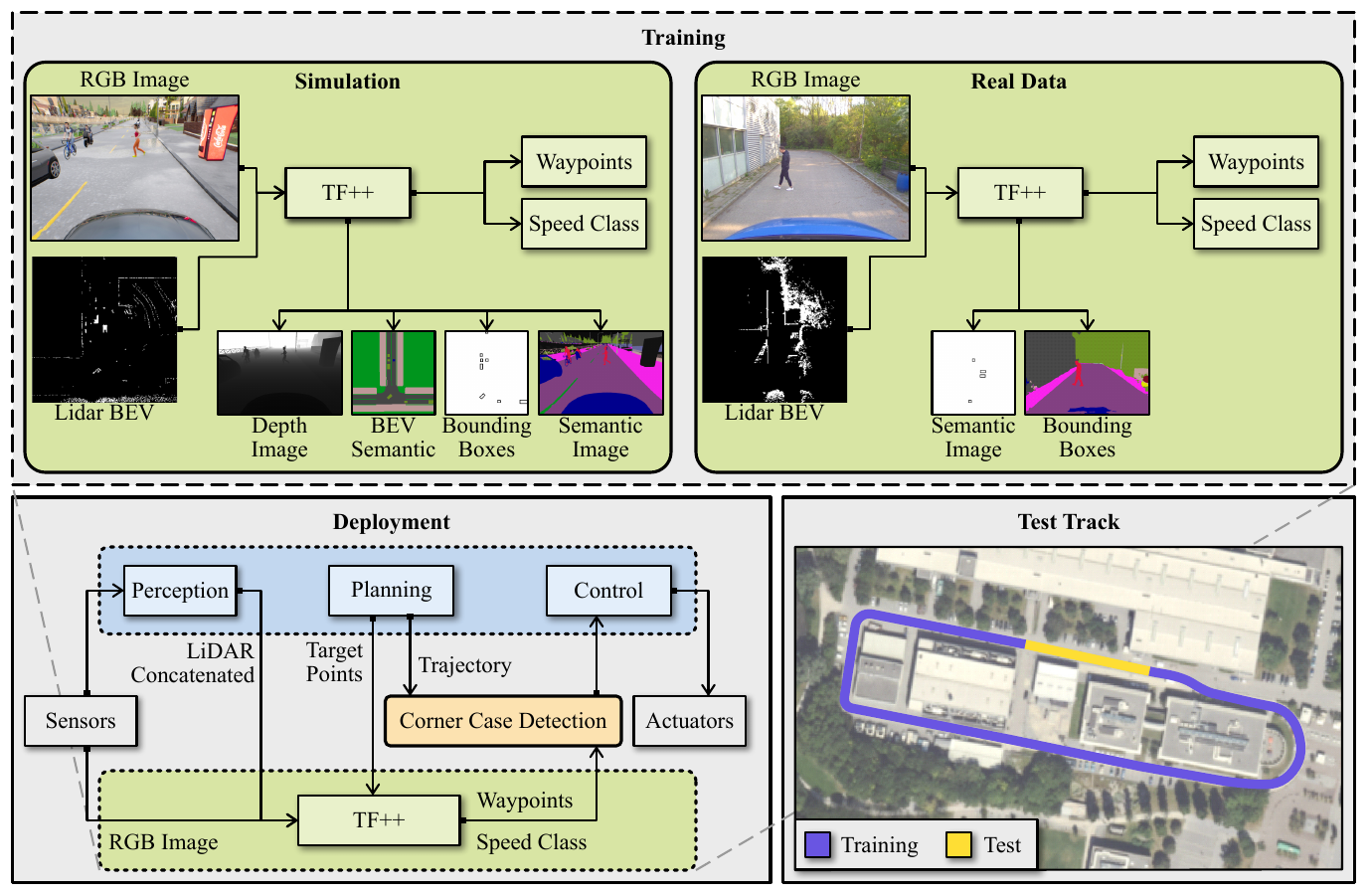}
\caption{Overview of the experiment design. The upper box illustrates the training process, including the inputs and outputs in the simulation and with real data. The box in the bottom left depicts the implementation of our method, showing only the modules relevant to the method. The test track is shown in the bottom right.}
\label{fig:experiment}
\end{figure*}%

By not requiring the network to fulfill the entire driving task, new possibilities for the output are opened up. The disadvantage of discontinuous speed and non-smooth trajectory can be circumvented thereby. For example, an end-to-end system may tend to predict an average speed from various modes instead of the target speed. This can be circumvented by the network not needing to predict an exact speed for the controller, but rather providing only an estimation of the possible speed for the current scenario. Secondly, the end-to-end system functions merely as an additional safeguard for the primary network, implying it does not necessarily need to operate in real-time with high frequency like the primary system. In this context, an interference time could be derived from the reflexes of the human safety driver, typically around \SI{1.2}{\second} \cite{johansson71} in unforeseen situations, enabling the possibility of larger network architectures. Since current end-to-end approaches often do not improve performance by incorporating control and frequently attach a rule-based controller to their network \cite{shao22, chitta22, chen22}, in our method, the end-to-end system covers the modules up to the planning stage. In the subsequent module, the output from the modular system, the longitudinal and lateral planned trajectory, is then compared with the output of the end-to-end system. A corner case is generally defined by a significant deviation between the two systems. In the event of significant deviations, a speed reduction, up to triggering a minimal risk maneuver, may be initiated accordingly. Various metrics can be utilized to measure the divergence; additionally, divergence over time can be considered in similarity to \cite{stocco20}. A concrete implementation is detailed in \cref{sec:experiment_design}.

    \section{Experiment Design}
\label{sec:experiment_design}
The following section presents a specific implementation of our approach, as well as the experimental setup to qualitatively evaluate it for corner case detection on a real vehicle. An overview of the experiment design is provided in \cref{fig:experiment}. An internal autonomous driving software based on Autoware Universe \cite{git_autoware} was used for the modular system, while the network architecture of \ac{tf++} \cite{jaeger23} served as the basis for the end-to-end system. This choice was primarily based on its decoupling of speed and waypoints and its strong performance in the CARLA Leaderboard Challenge \cite{dosovitskiy}. Real data and test drives were conducted using the EDGAR Research Vehicle \cite{karle2024edgar}. A Basler acA1920-50gc mono camera positioned at the front, along with two Ouster OS1-128 \ac{lidar} sensors, served as the primary sensors for environmental perception for both the modular and the end-to-end system.

\subsection{Training}
\label{subsec:training}
The end-to-end systems training occurred in two stages. Initially, extensive training was conducted in the simulation, followed by transfer learning using real data to bridge the sim-to-real gap. The training data was generated with the assistance of CARLA and the toolchain published in \cite{jaeger23}. The data were generated by a privileged rule-based driver, which the neural network was intended to imitate. The pre-trained \ac{tf++} served as the starting point, subsequently undergoing further training with the simulated sensor setup from the research vehicle. All available auxiliary outputs were utilized for training with simulation data: Depth Images, Semantic Images, \ac{bev} Semantic Images, and Bounding Boxes were included in the training process. 
The real data were recorded using the EDGAR vehicle with the assistance of a human driver. The modular system operated in shadow mode, meaning that its output was not communicated to the controller. To enhance training effectiveness, semantic images and bounding boxes were also utilized as auxiliary outputs for the real data. The semantic images were generated using \cite{zhao2017pspnet, mmseg2020}. The bounding boxes were extracted from the perception module of the modular stack running in shadow mode, the architecture used for object detection was based on \cite{yin2021centerbased}. The waypoints for training were derived from the vehicle's localization, with the waypoints generated based on the separation of longitudinal and lateral trajectories using spatial encoding rather than temporal encoding. The speed classes were partly determined manually and derived from the vehicle's accelerations. They were divided into four qualitative classes, each without a direct relation to the actual vehicle speed. The speed classes are listed in \cref{tab:speec-classes}.

\begin{table}[h]
\centering
\begin{tabular}{|c|c|p{5,5cm}|}
\hline
\textbf{Value} & \textbf{Name} & \textbf{Description} \\ \hline
3 & OK & No immediate danger \\ \hline
2 & Warning & Brakes may need to be applied in the near future \\ \hline
1 & Pedestrian & Braking is required due to the proximity of a pedestrian \\ \hline
0 & Brake & Brake immediately \\ \hline
\end{tabular}
\caption{Predicted speed classes.}
\label{tab:speec-classes}
\end{table}

\subsection{Deployment}
Given that the fundamental idea behind the presented concept is to leverage the advantages of both the modular and end-to-end approach, the concrete implementation also focused on capitalizing on synergies between the two. Both approaches receive identical sensor data for environment perception, with the end-to-end system receiving pre-processed, synchronized, and concatenated \ac{lidar} data from the modular system's perception. Additionally, the modular system utilizes data from supplementary sensors such as the vehicle's GNSS or odometry. As the end-to-end system requires behavioral input, target points are generated from the trajectory planned by the global planner within the modular system. The output of the end-to-end system and the planned trajectory from the modular planner are utilized for corner case detection.

\subsection{Corner Case Detection}
For corner case detection based on lateral differences, the planned trajectory of the modular system is compared with the predicted waypoints of the end-to-end system at each timestamp $t$. Let $lat_{mod_i}$ represent the lateral deviation at point $i$ along the trajectory as computed by the modular system, and $lat_{e2e_i}$ represent the lateral deviation at the same point $i$ along the trajectory as computed by end-to-end system. Then, the maximum difference in lateral deviations at point $i$ can be calculated as:

\begin{equation}
\ Lat_m(t) = \max_{i} |lat_{mod_i}(t) - lat_{e2e_i}(t)|
\end{equation}

where $Lat_m$ is the maximum difference in lateral deviations between corresponding points along the trajectories. In addition to that, the average deviation is calculated to mitigate the influence of individual outliers, ensuring they do not exert excessive impact.
\begin{equation}
\ Lat_{avg}(t) = \frac{1}{n} \sum_{i=1}^{n} |lat_{mod_i}(t) - lat_{e2e_i}(t)|
\end{equation}
The final metric is a combination of both formulas, utilizing weights:
\begin{equation}
\ Lat(t) = w_mLat_m(t) + w_{avg}Lat_{avg}(t)
\end{equation}
Since we only examine the results qualitatively in our work and do not have sufficient data, the weights could not be precisely determined and were both set to 1.

For longitudinal corner case detection, the speed from the planned trajectory of the modular system and the predicted speed class from the end-to-end system, which can also be interpreted as hazard classes, are utilized.
Let \( v \) represent the current speed, and let \( sc \) denote the speed class, where \( sc \in \{0, 1, 2, 3\} \). As already described in \cref{subsec:training},  0 indicates a high hazard requiring braking, and 3 indicates no hazard.

To monitor if the speed decreases when the speed class drops, we can define a corner case metric \( Long(t) \) as follows:

\begin{equation}
Long(t) = \begin{cases}
0 & \text{otherwise}, \\
1 & \delta sc(t) < 0 \land \delta v(t) \geq 0 \\
\end{cases}
\end{equation}

Here, \( \delta sc(t) \) represents the derivative of the speed class, while \( \delta v(t) \) refers to the derivative of the target velocity of the modular system. A corner case is detected if the end-to-end system predicts a more dangerous speed class and the modular system does not reduce its speed simultaneously. This indicates that the end-to-end network recognizes a more dangerous situation or demands deceleration, but the modular approach does not respond accordingly.

\subsection{Evaluation}
The evaluation took place on our test track, situated on private property yet subject to moderate traffic from other vehicles and pedestrians. A safety driver supervised the evaluation, intervening in critical situations as necessary. Simultaneously, output from the corner case detection was analyzed. The safety driver was able to see the planned trajectory of the primary system at all times and thus had the opportunity to intervene only in situations that the system would have solved incorrectly. Furthermore, corner cases were deliberately provoked by other road users on a section of the test route that had not previously been used in the training data. 
    \section{Results}
\label{sec:results}
The following chapter presents the results of our evaluation. Firstly, we showcase the results of the lateral trajectory evaluation, followed by the longitudinal trajectory evaluation. In both cases, one scenario was selected to represent the results.

\subsection{Lateral}

\begin{figure}[b!] 
    \centering 
    \includegraphics[width=\columnwidth]{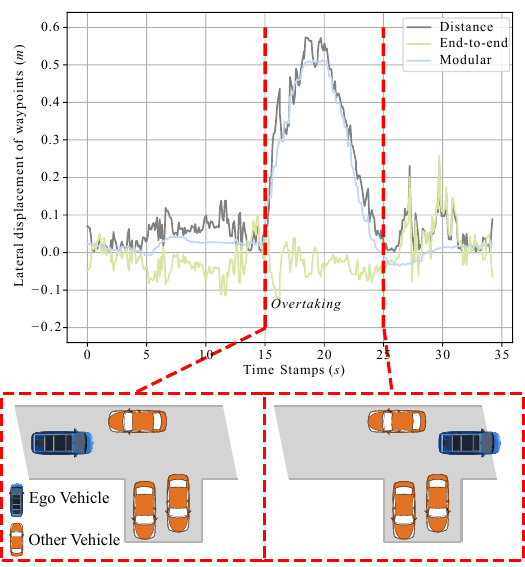}
    \caption{Visualization of the lateral part of the trajectory of both the modular system and the end-to-end system over time, during an overtaking maneuver, and the distance between the two. The corresponding scenario is additionally illustrated below.} 
    \label{fig:lat_res} 
\end{figure}

\cref{fig:lat_res} illustrates the lateral deviation of both the modular system and the end-to-end system in a specific scenario, with the distance between them also calculated. The scenario under analysis involves an overtaking maneuver, where the research vehicle passes a parked vehicle on the roadside. The figure indicates that while the modular system correctly initiates the overtaking maneuver, the end-to-end system consistently tries to remain in the middle of the lane, potentially resulting in a false positive corner case detection. Overall, the scenarios tested suggest that the difference in lateral trajectories between the two systems can be minimal, even during cornering, as long as there are no evasive maneuvers. However, during evasive maneuvers, it was observed that the end-to-end system consistently attempts to adhere to the global path, typically located in the center of the route, whereas the modular system often initiates overtaking maneuvers. Based on the available training data and test drives, no successful corner cases were detected through the analysis of lateral trajectories.

\begin{figure}[b!] 
    \centering 
    \includegraphics[width=\columnwidth]{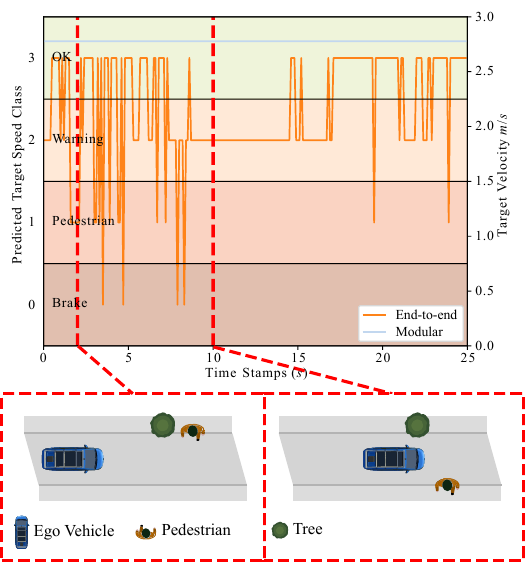}
    \caption{Visualization of the predicted speed class in orange and the target speed specified by the modular system. The corner case in which the safety driver has braked is shown with the red dashed lines. The corresponding scenario is additionally illustrated below.} 
    \label{fig:res_long} 
\end{figure}

\subsection{Longitudinal}

\cref{fig:res_long} illustrates an example of the result from the longitudinal evaluation for a selected scenario. The orange curve represents the predicted speed class, while the target speed of the modular system is depicted in blue. In this scenario, the research vehicle autonomously traveled along a straight line. Within the area highlighted in red, the safety driver noticed a malfunction in the modular system and intervened. During this period, a pedestrian entered the road from a slightly obscured angle, and the modular system failed to react correctly to avoid a collision. The fault could not be attributed to a single module and presumably resulted from a combination of factors across different modules. Within the red zone, it is evident that the end-to-end system correctly identified a more hazardous situation and therefore advocated for slower driving. The disparity in behavior compared to the modular system could be effectively utilized for corner case detection. Overall, the longitudinal evaluation of the driving test scenarios showed promising results, but there were also instances of undetected corner cases or false positives.

    \section{Discussion}
\label{sec:discussion}

The lateral results yielded limited success in corner case detection. This was primarily attributed to the end-to-end approach's inability to navigate around obstacles. The underlying reason is that the expert driver, whose recorded simulation data was used for imitation learning, was incapable of evading objects on the road, and such scenarios were absent from the CARLA Leaderboard. Nevertheless, it was observed that the lateral disparity between the end-to-end and modular systems could be minimized in real-world settings. Conversely, longitudinal results indicated the potential success of the proposed corner case detection. The end-to-end system, capable of identifying hazardous situations and responding appropriately, could have averted a collision in the mentioned case. Generally, the output of hazard classes emerged as a dependable method compared to specific target speeds. However, all results were significantly affected by pronounced noise, mainly stemming from the extensive reliance on simulation data for training the end-to-end system, which struggles to fully adapt to real-world conditions due to the simulation-to-reality gap. Additionally, the lack of diversity in the CARLA Leaderboard scenarios, particularly the underrepresentation of pedestrian-related scenarios, posed another challenge. Furthermore, the qualitative results did not indicate the number of false positives and false negatives in the corner case detection. False positives can occur when a situation can be resolved in multiple ways, such as in an evasive maneuver where the modular approach might deviate to the left while the end-to-end approach possibly deviates to the right. These false positives might be partially mitigated by the fact that the corner case analysis is conducted over time, and after an initial significant difference, the end-to-end system may eventually align with the behavior of the modular system. However, an additional rule-based or data-driven approach might also be necessary to account for ambiguous situations, such as evasive maneuvers or merging maneuvers, in corner case detection. False negatives can occur if both systems make the same incorrect decision.

    \section{Conclusion}
\label{sec:conclusion}

Online corner case detection is crucial for ensuring safe autonomous driving. This paper introduces a concept a for online corner case detection that merges the traditional modular approach with the end-to-end approach to leverage the strengths of both methods. The proposed approach was implemented using an open-source end-to-end driving architecture as a demonstration and trained using a combination of simulation and real-world data. Longitudinal prediction was facilitated through the utilization of speed classes, which offer insights into current hazard assessment rather than prescribing actual driving speeds. Qualitative findings demonstrate the system's potential to effectively identify corner cases in real-world driving scenarios.

In the future, the concept should undergo quantitative evaluation, enabling comparison between true positive and false positive detections through a higher number of trials. Furthermore, real-world data for network training should be substantially expanded. The paper has already demonstrated that employing the end-to-end system as a secondary system opens up different possibilities regarding input, output, and runtime. These architectural components warrant further investigation to determine a suitable network architecture for end-to-end systems as secondary driving systems for online corner case detection.
     \section*{Acknowledgment}
Gemb Kaljavesi was the main developer of the presented work and main contributer of this
paper. Xiyan Su supported in performing the training and evaluation of the proposed approach and revised the paper critically. Frank Diermeyer made contributions to the conception of the research project and revised the paper
critically. The authors want to thank Florian Pfab, Domagoj Majstorovic, Niklas Krauß, Tobias Kerbl and Dominik Kulmer for their support during the experiment phase, and Bernhard Jäger for his assistance with using the data generation pipeline. 
    \begin{acronym}

    \acro{av}[AV]{Automated Vehicle}
    \acro{who}[WHO]{World Health Organization}
    \acro{mpc}[MPC]{Model Predictive Control}
    \acro{pid}[PID]{Proportional-Integral-Derivative}
    \acro{hd}[HD]{High Definition}
    \acroplural{av}[AVs]{Automated Vehicles}
    \acro{ros}[ROS]{Robot Operating System}
    \acro{hf}[HF]{High-Fidelity}
    \acro{lidar}[LiDAR]{Light Detecting and Ranging}
    \acro{tf++}[TF++]{TransFuser++}
    \acro{bev}[BEV]{Bird's Eye View}

\end{acronym}

    \bibliographystyle{IEEEtran}
    \bibliography{main}

\begin{thebibliography}{10}
\providecommand{\url}[1]{#1}
\csname url@samestyle\endcsname
\providecommand{\newblock}{\relax}
\providecommand{\bibinfo}[2]{#2}
\providecommand{\BIBentrySTDinterwordspacing}{\spaceskip=0pt\relax}
\providecommand{\BIBentryALTinterwordstretchfactor}{4}
\providecommand{\BIBentryALTinterwordspacing}{\spaceskip=\fontdimen2\font plus
\BIBentryALTinterwordstretchfactor\fontdimen3\font minus
  \fontdimen4\font\relax}
\providecommand{\BIBforeignlanguage}[2]{{%
\expandafter\ifx\csname l@#1\endcsname\relax
\typeout{** WARNING: IEEEtran.bst: No hyphenation pattern has been}%
\typeout{** loaded for the language `#1'. Using the pattern for}%
\typeout{** the default language instead.}%
\else
\language=\csname l@#1\endcsname
\fi
#2}}
\providecommand{\BIBdecl}{\relax}
\BIBdecl

\bibitem{tafidis19}
P.~Tafidis, A.~Pirdavani, T.~Brijs, and H.~Farah, ``Can {{Automated Vehicles
  Improve Cyclist Safety}} in {{Urban Areas}}?'' \emph{Safety}, vol.~5, no.~3,
  p.~57, Aug. 2019.

\bibitem{morando18}
M.~M. Morando, Q.~Tian, L.~T. Truong, and H.~L. Vu, ``Studying the {{Safety
  Impact}} of {{Autonomous Vehicles Using Simulation-Based Surrogate Safety
  Measures}},'' \emph{Journal of Advanced Transportation}, vol. 2018, pp.
  1--11, 2018.

\bibitem{combs19}
T.~S. Combs, L.~S. Sandt, M.~P. Clamann, and N.~C. McDonald, ``Automated
  {{Vehicles}} and {{Pedestrian Safety}}: {{Exploring}} the {{Promise}} and
  {{Limits}} of {{Pedestrian Detection}},'' \emph{American Journal of
  Preventive Medicine}, vol.~56, no.~1, pp. 1--7, Jan. 2019.

\bibitem{utriainen20}
R.~Utriainen and M.~P{\"o}ll{\"a}nen, ``Prioritizing {{Safety}} or {{Traffic
  Flow}}? {{Qualitative Study}} on {{Highly Automated Vehicles}}' {{Potential}}
  to {{Prevent Pedestrian Crashes}} with {{Two Different Ambitions}},''
  \emph{Sustainability}, vol.~12, no.~8, p. 3206, Apr. 2020.

\bibitem{tafidis22}
P.~Tafidis, H.~Farah, T.~Brijs, and A.~Pirdavani, ``Safety implications of
  higher levels of automated vehicles: A scoping review,'' \emph{Transport
  Reviews}, vol.~42, no.~2, pp. 245--267, Mar. 2022.

\bibitem{yurtsever20}
E.~Yurtsever, J.~Lambert, A.~Carballo, and K.~Takeda, ``A {{Survey}} of
  {{Autonomous Driving}}: {{Common Practices}} and {{Emerging Technologies}},''
  \emph{IEEE Access}, vol.~8, pp. 58\,443--58\,469, 2020.

\bibitem{tampuu22}
\BIBentryALTinterwordspacing
A.~Tampuu, T.~Matiisen, M.~Semikin, D.~Fishman, and N.~Muhammad, ``A {{Survey}}
  of {{End-to-End Driving}}: {{Architectures}} and {{Training Methods}},''
  vol.~33, no.~4, pp. 1364--1384. [Online]. Available:
  \url{https://ieeexplore.ieee.org/document/9310544/}
\BIBentrySTDinterwordspacing

\bibitem{chib23}
P.~S. Chib and P.~Singh, ``Recent advancements in end-to-end autonomous driving
  using deep learning: A survey,'' \emph{IEEE Transactions on Intelligent
  Vehicles}, vol.~9, no.~1, pp. 103--118, 2024.

\bibitem{dosovitskiy}
A.~Dosovitskiy, ``{{CARLA}}: {{An Open Urban Driving Simulator}}.''

\bibitem{shao23a}
\BIBentryALTinterwordspacing
H.~Shao, L.~Wang, R.~Chen, S.~L. Waslander, H.~Li, and Y.~Liu, ``Reasonnet:
  End-to-end driving with temporal and global reasoning,'' in \emph{2023
  IEEE/CVF Conference on Computer Vision and Pattern Recognition (CVPR)}.\hskip
  1em plus 0.5em minus 0.4em\relax Los Alamitos, CA, USA: IEEE Computer
  Society, jun 2023, pp. 13\,723--13\,733. [Online]. Available:
  \url{https://doi.ieeecomputersociety.org/10.1109/CVPR52729.2023.01319}
\BIBentrySTDinterwordspacing

\bibitem{jain21}
A.~Jain, L.~Del~Pero, H.~Grimmett, and P.~Ondruska, ``Autonomy 2.0: {{Why}} is
  self-driving always 5 years away?'' Aug. 2021.

\bibitem{schlatow17}
J.~Schlatow, M.~Moostl, R.~Ernst, M.~Nolte, I.~Jatzkowski, M.~Maurer,
  C.~Herber, and A.~Herkersdorf, ``Self-awareness in autonomous automotive
  systems,'' in \emph{Design, {{Automation}} \& {{Test}} in {{Europe
  Conference}} \& {{Exhibition}} ({{DATE}}), 2017}.\hskip 1em plus 0.5em minus
  0.4em\relax {Lausanne, Switzerland}: {IEEE}, Mar. 2017, pp. 1050--1055.

\bibitem{bogdoll22}
\BIBentryALTinterwordspacing
D.~Bogdoll, M.~Nitsche, and J.~M. Zollner, ``Anomaly {{Detection}} in
  {{Autonomous Driving}}: {{A Survey}},'' in \emph{2022 {{IEEE}}/{{CVF
  Conference}} on {{Computer Vision}} and {{Pattern Recognition Workshops}}
  ({{CVPRW}})}.\hskip 1em plus 0.5em minus 0.4em\relax IEEE, pp. 4487--4498.
  [Online]. Available: \url{https://ieeexplore.ieee.org/document/9857500/}
\BIBentrySTDinterwordspacing

\bibitem{git_autoware}
\BIBentryALTinterwordspacing
``{Autoware}.'' [Online]. Available:
  \url{https://github.com/autowarefoundation/ autoware}
\BIBentrySTDinterwordspacing

\bibitem{git_apollo}
\BIBentryALTinterwordspacing
``{Apollo}.'' [Online]. Available: \url{https://github.com/ApolloAuto/apollo}
\BIBentrySTDinterwordspacing

\bibitem{betz19}
\BIBentryALTinterwordspacing
{Betz}, {Heilmeier}, {Wischnewski}, {Stahl}, and {Lienkamp}, ``Autonomous
  {{Driving}}—{{A Crash Explained}} in {{Detail}},'' vol.~9, no.~23, p. 5126.
  [Online]. Available: \url{https://www.mdpi.com/2076-3417/9/23/5126}
\BIBentrySTDinterwordspacing

\bibitem{plaza}
L.~Plaza, ``Collision {{Between Vehicle Controlled}} by {{Developmental
  Automated Driving System}} and {{Pedestrian}}, {{Tempe}}, {{Arizona}},
  {{March}} 18, 2018.''

\bibitem{stahl}
T.~N. Stahl, ``Safeguarding {{Complex}} and {{Learning-Based Automated Driving
  Functions}} via {{Online Verification}},'' p. 220.

\bibitem{gruyer17}
\BIBentryALTinterwordspacing
D.~Gruyer, V.~Magnier, K.~Hamdi, L.~Claussmann, O.~Orfila, and
  A.~Rakotonirainy, ``Perception, information processing and modeling:
  {{Critical}} stages for autonomous driving applications,'' vol.~44, pp.
  323--341. [Online]. Available:
  \url{https://linkinghub.elsevier.com/retrieve/pii/S136757881730113X}
\BIBentrySTDinterwordspacing

\bibitem{shao22}
\BIBentryALTinterwordspacing
H.~Shao, L.~Wang, R.~Chen, H.~Li, and Y.~Liu. Safety-{{Enhanced Autonomous
  Driving Using Interpretable Sensor Fusion Transformer}}. [Online]. Available:
  \url{http://arxiv.org/abs/2207.14024}
\BIBentrySTDinterwordspacing

\bibitem{chen22}
\BIBentryALTinterwordspacing
D.~Chen and P.~Krahenbuhl, ``Learning from {{All Vehicles}},'' in \emph{2022
  {{IEEE}}/{{CVF Conference}} on {{Computer Vision}} and {{Pattern
  Recognition}} ({{CVPR}})}.\hskip 1em plus 0.5em minus 0.4em\relax IEEE, pp.
  17\,201--17\,210. [Online]. Available:
  \url{https://ieeexplore.ieee.org/document/9879043/}
\BIBentrySTDinterwordspacing

\bibitem{chen24}
L.~Chen, P.~Wu, K.~Chitta, B.~Jaeger, A.~Geiger, and H.~Li, ``End-to-end
  autonomous driving: Challenges and frontiers,'' \emph{IEEE Transactions on
  Pattern Analysis and Machine Intelligence}, pp. 1--20, 2024.

\bibitem{chen20}
J.~Chen, S.~E. Li, and M.~Tomizuka, ``Interpretable end-to-end urban autonomous
  driving with latent deep reinforcement learning,'' \emph{IEEE Transactions on
  Intelligent Transportation Systems}, vol.~23, no.~6, pp. 5068--5078, 2022.

\bibitem{hecker2019learning}
S.~Hecker, D.~Dai, and L.~V. Gool, ``Learning accurate, comfortable and
  human-like driving,'' 2019.

\bibitem{stahl21a}
\BIBentryALTinterwordspacing
T.~Stahl and F.~Diermeyer, ``Online {{Verification Enabling Approval}} of
  {{Driving Functions}}—{{Implementation}} for a {{Planner}} of an
  {{Autonomous Race Vehicle}},'' vol.~2, pp. 97--110. [Online]. Available:
  \url{https://ieeexplore.ieee.org/document/9424710/}
\BIBentrySTDinterwordspacing

\bibitem{breitenstein20}
J.~Breitenstein, J.-A. Termohlen, D.~Lipinski, and T.~Fingscheidt,
  ``Systematization of {{Corner Cases}} for {{Visual Perception}} in
  {{Automated Driving}},'' in \emph{2020 {{IEEE Intelligent Vehicles
  Symposium}} ({{IV}})}.\hskip 1em plus 0.5em minus 0.4em\relax {Las Vegas, NV,
  USA}: {IEEE}, Oct. 2020, pp. 1257--1264.

\bibitem{bolte19}
J.-A. Bolte, A.~Bar, D.~Lipinski, and T.~Fingscheidt, ``Towards {{Corner Case
  Detection}} for {{Autonomous Driving}},'' in \emph{2019 {{IEEE Intelligent
  Vehicles Symposium}} ({{IV}})}.\hskip 1em plus 0.5em minus 0.4em\relax
  {Paris, France}: {IEEE}, Jun. 2019, pp. 438--445.

\bibitem{heidecker21}
F.~Heidecker, J.~Breitenstein, K.~R{\"o}sch, J.~L{\"o}hdefink, M.~Bieshaar,
  C.~Stiller, T.~Fingscheidt, and B.~Sick, ``An {{Application-Driven
  Conceptualization}} of {{Corner Cases}} for {{Perception}} in {{Highly
  Automated Driving}},'' in \emph{2021 {{IEEE Intelligent Vehicles Symposium}}
  ({{IV}})}, Jul. 2021, pp. 644--651.

\bibitem{isopas_21448_iso_2022}
{International Organization for Standardization}, ``{ISO/PAS 21448: Road
  vehicles - Safety of the intended functionality},'' 2022.

\bibitem{stocco20}
\BIBentryALTinterwordspacing
A.~Stocco, M.~Weiss, M.~Calzana, and P.~Tonella, ``Misbehaviour prediction for
  autonomous driving systems,'' in \emph{Proceedings of the {{ACM}}/{{IEEE}}
  42nd {{International Conference}} on {{Software Engineering}}}.\hskip 1em
  plus 0.5em minus 0.4em\relax ACM, pp. 359--371. [Online]. Available:
  \url{https://dl.acm.org/doi/10.1145/3377811.3380353}
\BIBentrySTDinterwordspacing

\bibitem{hussain22}
\BIBentryALTinterwordspacing
M.~Hussain, N.~Ali, and J.-E. Hong, ``{{DeepGuard}}: A framework for
  safeguarding autonomous driving systems from inconsistent behaviour,''
  vol.~29, no.~1, p.~1. [Online]. Available:
  \url{https://link.springer.com/10.1007/s10515-021-00310-0}
\BIBentrySTDinterwordspacing

\bibitem{kuhn20}
\BIBentryALTinterwordspacing
C.~B. Kuhn, M.~Hofbauer, G.~Petrovic, and E.~Steinbach, ``Introspective {{Black
  Box Failure Prediction}} for {{Autonomous Driving}},'' in \emph{2020 {{IEEE
  Intelligent Vehicles Symposium}} ({{IV}})}.\hskip 1em plus 0.5em minus
  0.4em\relax IEEE, pp. 1907--1913. [Online]. Available:
  \url{https://ieeexplore.ieee.org/document/9304844/}
\BIBentrySTDinterwordspacing

\bibitem{kuhn22}
\BIBentryALTinterwordspacing
C.~B. Kuhn, M.~Hofbauer, G.~Petrovic, and E.~Steinbach, ``Introspective
  {{Failure Prediction}} for {{Autonomous Driving Using Late Fusion}} of
  {{State}} and {{Camera Information}},'' vol.~23, no.~5, pp. 4445--4459.
  [Online]. Available: \url{https://ieeexplore.ieee.org/document/9310689/}
\BIBentrySTDinterwordspacing

\bibitem{fridman18}
\BIBentryALTinterwordspacing
L.~Fridman, L.~Ding, B.~Jenik, and B.~Reimer. Arguing {{Machines}}: {{Human
  Supervision}} of {{Black Box AI Systems That Make Life-Critical Decisions}}.
  [Online]. Available: \url{http://arxiv.org/abs/1710.04459}
\BIBentrySTDinterwordspacing

\bibitem{hanselaar24b}
\BIBentryALTinterwordspacing
C.~A.~J. Hanselaar, E.~Silvas, A.~Terechko, and W.~P. M.~H. Heemels, ``The
  {{Safety Shell}}: An {{Architecture}} to {{Handle Functional
  Insufficiencies}} in {{Automated Driving}},'' pp. 1--19. [Online]. Available:
  \url{http://arxiv.org/abs/2311.08413}
\BIBentrySTDinterwordspacing

\bibitem{johansson71}
G.~Johansson and K.~Rumar, ``Drivers' {{Brake Reaction Times}},'' \emph{Human
  Factors: The Journal of the Human Factors and Ergonomics Society}, vol.~13,
  no.~1, pp. 23--27, Feb. 1971.

\bibitem{chitta22}
K.~Chitta, A.~Prakash, B.~Jaeger, Z.~Yu, K.~Renz, and A.~Geiger,
  ``{{TransFuser}}: {{Imitation}} with {{Transformer-Based Sensor Fusion}} for
  {{Autonomous Driving}},'' May 2022.

\bibitem{jaeger23}
B.~Jaeger, K.~Chitta, and A.~Geiger, ``Hidden {{Biases}} of {{End-to-End
  Driving Models}},'' Aug. 2023.

\bibitem{karle2024edgar}
P.~Karle, T.~Betz, M.~Bosk, F.~Fent, N.~Gehrke, M.~Geisslinger, L.~Gressenbuch,
  P.~Hafemann, S.~Huber, M.~Hübner, S.~Huch, G.~Kaljavesi, T.~Kerbl,
  D.~Kulmer, T.~Mascetta, S.~Maierhofer, F.~Pfab, F.~Rezabek, E.~Rivera,
  S.~Sagmeister, L.~Seidlitz, F.~Sauerbeck, I.~Tahiraj, R.~Trauth, N.~Uhlemann,
  G.~Würsching, B.~Zarrouki, M.~Althoff, J.~Betz, K.~Bengler, G.~Carle,
  F.~Diermeyer, J.~Ott, and M.~Lienkamp, ``Edgar: An autonomous driving
  research platform -- from feature development to real-world application,''
  2024.

\bibitem{zhao2017pspnet}
H.~Zhao, J.~Shi, X.~Qi, X.~Wang, and J.~Jia, ``Pyramid scene parsing network,''
  in \emph{CVPR}, 2017.

\bibitem{mmseg2020}
M.~Contributors, ``{MMSegmentation}: Openmmlab semantic segmentation toolbox
  and benchmark,'' \url{https://github.com/open-mmlab/mmsegmentation}, 2020.

\bibitem{yin2021centerbased}
T.~Yin, X.~Zhou, and P.~Krähenbühl, ``Center-based 3d object detection and
  tracking,'' 2021.

\end{thebibliography}

\end{document}